\pdfoutput=1
\documentclass[11pt]{article}

\usepackage[]{emnlp2021}

\usepackage{times}
\usepackage{latexsym}
\usepackage{booktabs}
\usepackage{graphicx}
\usepackage{tikz}
\usetikzlibrary{positioning,shapes,decorations,automata}
\usepackage{pifont}
\usepackage{caption}
\usepackage{subcaption}

\usepackage[T1]{fontenc}
\usepackage[utf8]{inputenc}
\usepackage[scaled=0.8]{beramono}

\usepackage{microtype}

\title{Cross-document Event Identity via Dense Annotation}
\author{Adithya Pratapa, Zhengzhong Liu, Kimihiro Hasegawa, \\ {\bf Linwei Li, Yukari Yamakawa, Shikun Zhang, Teruko Mitamura} \\
Language Technologies Institute, Carnegie Mellon University \\
\texttt{\{vpratapa,zhengzhl,kimihiro,linweil,yukariy,shikunz,teruko\}@andrew.cmu.edu}
}

\newcommand{\cmark}{\ding{52}}%

\begin{document}
\maketitle
\begin{abstract}
In this paper, we study the identity of textual events from different documents. While the complex nature of event identity is previously studied \cite{hovy-etal-2013-events}, the case of events across documents is unclear. Prior work on cross-document event coreference has two main drawbacks. First, they restrict the annotations to a limited set of event types. Second, they insufficiently tackle the concept of event identity.  Such annotation setup reduces the pool of event mentions and prevents one from considering the possibility of quasi-identity relations. We propose a dense annotation approach for cross-document event coreference, comprising a rich source of event mentions and a dense annotation effort between related document pairs. To this end, we design a new annotation workflow with careful quality control and an easy-to-use annotation interface. In addition to the links, we further collect overlapping event contexts, including time, location, and participants, to shed some light on the relation between identity decisions and context. We present an open-access dataset for cross-document event coreference, CDEC-WN, collected from English Wikinews and open-source our annotation toolkit to encourage further research on cross-document tasks.\footnote{Data and code are available at \url{https://github.com/adithya7/cdec-wikinews}.}
\end{abstract}

\section{Introduction}
\label{sec:introduction}

Coreference resolution is the task of identifying events (or entities) that refer to the same underlying activity (or objects). Accurately resolving coreference is a prerequisite for many NLP tasks, such as question answering, summarization, and dialogue understanding. For instance, to get a holistic view of an ongoing natural disaster, we need to aggregate information from various sources (newswire, social media, public communication, etc.) over an extended period. Often this requires resolving coreference between mentions across documents.\footnote{A mention is a linguistic expression in text that denotes a specific instance of an event.}

\begin{figure}[t]
\centering
\resizebox{0.48\textwidth}{!}{
\begin{tikzpicture}[
rectnode/.style={rectangle, thick, align=center, minimum size=2.5cm},
]
%Nodes
\node[rectnode] (a) [text width=6cm] {(October 23, 2010) Nearly \emph{200 people} are confirmed dead and approximately \emph{2600} are ill in a central Haitian cholera \textbf{outbreak}.};
\node[rectnode] (b) [below left=1cm and -2.5cm of a,text width=3.5cm] {(October 26, 2010) At least \emph{259 people} are dead and over \emph{3000 people} have been infected in the Haitian cholera \textbf{outbreak}.};
\node[rectnode] (c) [below right=1cm and -2.5cm of a,text width=3.5cm] {(October 28, 2010) The Haitian cholera \textbf{outbreak} has killed \emph{292 people} and infected over \emph{4000}, according to the Haitian government.};

%Lines
\draw[<->,draw=blue,very thick] (b.east) -- (c.west) node[midway,below] {\cmark};
\draw[<->,draw=blue,very thick] ([xshift=-0.1cm]a.south) -- ([xshift=0.1cm]b.north) node[midway,above left] {\cmark};
\draw[<->,draw=blue,very thick] ([xshift=0.1cm]a.south) -- ([xshift=-0.1cm]c.north) node[midway,above right] {\cmark};

\end{tikzpicture}}

\caption{An illustration of the quasi-identity nature of events. The event [Haitian cholera] `outbreak' is expressed by instances with varying counts of infections and deaths. The identity of this event continuously evolves over space and time, attributed to a new type of quasi-identity, spatiotemporal continuity.}
\label{fig:spatiotemporal_continuity_example}
\end{figure}
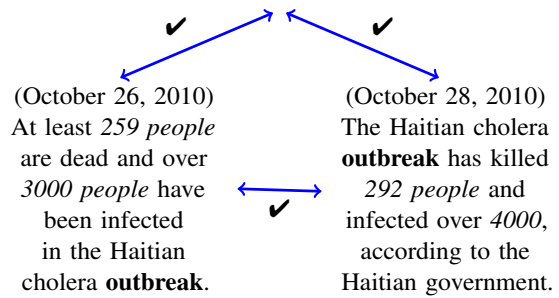

\citet{Recasens2011IdentityNA} defines coreference as ``identity of reference''. Therefore, modeling event coreference requires understanding the extent of the shared identity between event mentions. Numerous factors determine this identity, including the semantics of the event mention, its arguments, and the document context. Resolving coreference across documents is more challenging, as it requires modeling identity over a much longer context. To this end, we identify two major issues with existing cross-document event coreference (CDEC) datasets that limit the progress on this task. First, many prior datasets often annotate coreference \emph{only on a restricted set of event types}, limiting the coverage of mentions in the dataset. Second, many datasets and models \emph{insufficiently tackle the concept of event identity}. As highlighted by \citet{hovy-etal-2013-events}, the decision of whether two mentions refer to the same event is often non-trivial. Occasionally, event mentions only share a partial identity (\textit{quasi-identity}). In this work, we present a new dataset for CDEC that attempts to overcome both issues.

Earlier efforts on CDEC dataset collection were limited to specific pre-defined event types, restricting the scope of event mentions that could be studied. In this work, we instead annotate mentions of all types, i.e., open-domain events \cite{araki-mitamura-2018-open}, and provide a \emph{dense annotation} \cite{cassidy-etal-2014-annotation} by checking for coreference relationship between every mention pair in all underlying document pairs. We compile documents from the publicly available English Wikinews.\footnote{\url{https://en.wikinews.org/}} To facilitate our goal of dense annotation of mentions and their coreference, we develop and release a new easy-to-use annotation tool that allows linking text spans across documents. We crowdsource coreference annotations on Mechanical Turk.\footnote{\url{https://www.mturk.com/}}

Prior work has attributed the quasi-identity behavior of events to two specific phenomena, membership and subevent \cite{hovy-etal-2013-events}. However, its implications in cross-document settings remain unclear. In this work, we specifically focus on a cross-document setup. As highlighted by \citet{recasens-etal-2012-annotating}, a direct annotation of quasi-identity relations is hard because annotators might not be familiar with the phenomenon. Therefore, we propose a new annotation workflow that allows for easy determination of quasi-identity links. To this end, we collect evidence for time, location, and participant(s) overlap between corefering mentions. We also collect information regarding any potential inclusion relationship between the mention pair.

Our workflow allowed us to empirically identify a new type of quasi-identity, \emph{spatiotemporal continuity}, in addition to the existing types defined by~\citet{hovy-etal-2013-events}. \autoref{fig:spatiotemporal_continuity_example} illustrates this phenomenon using the case of [Haitian cholera] outbreak. The event gradually evolves over space and time, leading to cases of partial coreference. Additionally, traditional coreference annotations cluster mentions together. However, this methodology can be misleading when dealing with cases of quasi-identity (see \S\ref{sec:quasi_identity_analysis}). To overcome this limitation, we frame our annotation task as a (cross-document) mention pair linking. The proposed task simplifies the annotation process by avoiding merging quasi-identical mentions into a single cluster.

 The main contributions of our work can be summarized as follows,

\begin{itemize}
    \item We present an empirical study of the quasi-identity of events in the context of CDEC. In addition to providing evidence for previously studied types of quasi-identity (membership, subevent), we identify a novel type relating to the spatiotemporal continuity of events.
    \item We release a densely annotated CDEC dataset, CDEC-WN, spanning 198 document pairs across 55 subtopics from English Wikinews. The dataset is available under an open license. To serve as a benchmark for future work, we provide two baselines, lemma-match, and a BERT-based cross-encoder.
    \item To efficiently collect evidence for quasi-identity, we develop a novel annotation workflow built upon a custom-designed annotation tool. We deploy the workflow to crowdsource CDEC annotations from Mechanical Turk. 
\end{itemize}

In the upcoming sections, we first position our work within the existing CDEC literature (\S\ref{sec:related_work}). We then describe our methodology for preparing the source corpus (\S\ref{sec:corpus_preparation}), and our crowdsourcing setup for collecting coreference annotations on this corpus (\S\ref{sec:mturk_setup}). In \S\ref{sec:quasi_identity_analysis}, we present a study of quasi-identity of events in our dataset. Finally, in \S\ref{sec:baselines}, we present two baselines models for the proposed dataset.

\section{Related Work}\label{sec:related_work}

\paragraph{Event Coreference: } Widely studied in the literature, with datasets curated for both within and cross-document tasks. ACE 2005~\cite{walker2006ace}, OntoNotes~\cite{weischedel2013ontonotes}, and TAC-KBP~\cite{mitamura2017events} are commonly used benchmarks for within-document coreference. For cross-document coreference, ECB+~\cite{cybulska-vossen-2014-using} is a widely popular benchmark and is an extended version of the original ECB dataset~\cite{bejan-harabagiu-2008-linguistic}. ECB+ suffers from a major limitation with coreference annotations restricted to only the first few sentences in the documents. However, CDEC is a long-range phenomenon, and there is a need for more densely annotated datasets.

Many other datasets have since been curated for the task of CDEC. Some related works include, MEANTIME \cite{minard-etal-2016-meantime}, Event hoppers \cite{song-etal-2018-cross}, Gun Violence Corpus (GVC) \cite{vossen-etal-2018-dont}, Football Coreference Corpus (FCC) \cite{DBLP:conf/ecir/BugertRBDG20}, and Wikipedia Event Coreference (WEC) \cite{eirew-etal-2021-wec}. However, most CDEC systems are still evaluated primarily on ECB+. Additionally, all of these datasets do not account for the quasi-identity nature of events.

Though compiled from Wikinews, CDEC annotations in the MEANTIME corpus were limited to events with participants from a pre-defined list of 44 seed entities. While the FCC corpus was also crowdsourced, the annotation unit was an entire sentence instead of a single event mention. WEC corpus uses hyperlinks from Wikipedia but primarily handles referential events. In this work, we use open-domain events and treat an event mention as our annotation unit. We collect coreference links across all the mention pairs from all the underlying document pairs.

\paragraph{Event Identity: } \citet{Recasens2011IdentityNA} postulated entity coreference as a continuum, with identity, non-identity and near-identity relations. In a follow-up work \cite{recasens-etal-2012-annotating}, they identify near-identity relations using the disagreement between annotators. They say subjects are not fully aware of the near-identity behavior, therefore making direct annotation collection hard. The continuum idea has since extended to events \cite{hovy-etal-2013-events}. Determining if two event mentions are identical is not a trivial decision. It depends on the arguments of the mentions (often underspecified in the local context), the semantics of the mention, and the document context. In this work, we are specifically interested in cross-document coreference. \citet{wright-bettner-etal-2019-cross} studied the impact of the subevent relationship on quasi-identity, but a more general annotation framework is missing. Accurately capturing event identity is critical to CDEC dataset construction and the subsequent modeling. Therefore, we qualitatively study this phenomenon by collecting supplementary information with each coreference link.

\section{Corpus Preparation}\label{sec:corpus_preparation}

In our goal of curating a CDEC dataset, we first needed to identify documents that exhibit cross-document coreference. We now describe our document collection process and our methodology for annotating event mentions in these documents.

\paragraph{Document Selection:} To facilitate the redistribution of the documents under an open license, we prioritized collecting the documents from publicly available news sources. We chose Wikinews for three key reasons. First, the news articles were sourced from trusted news outlets and reported impartially. Second, these articles are available under an open license (CC BY 2.5), allowing easy redistribution. Finally, each article is human-labeled with categories (e.g., Disaster and accidents, Health, Sports, etc.),\footnote{\url{https://en.wikinews.org/wiki/Wikinews:Categories_and_topic_pages}} as we describe later, this meta-information plays a significant role in our dataset collection. We use the July 1st, 2020 dump of English Wikinews, which contains a total of 21k titles (or articles/documents). These news articles are timestamped from November 2004 to July 2020. Annotating coreference between every document pair in Wikinews is infeasible. Therefore, we first identify groups of related news articles. Articles within a given group usually describe a part of a developing news story or storyline.

\paragraph{Identifying Storylines:} To identify these latent storylines, we first construct an undirected Wikinews graph ($W$) with articles as nodes and add an edge between two nodes if one is mentioned under the ``Related News'' section in the other. We then identify cliques ($C_W$) (i.e., fully connected sub-graphs) in the Wikinews graph, which constitute our potential set of storylines. While the articles within each clique are related, we also want to minimize the relatedness of articles across cliques. Therefore, we construct a new graph ($M$), where each clique ($\in C_W$) is a node, and an edge is added between two nodes if the two cliques are not disjoint or if any two articles in the two cliques share an edge in the Wikinews graph ($W$). Finally, we extract maximal independent sets from $M$ that correspond to separate storylines. Among the multiple feasible maximal independent sets, we optimize for maximum overlap in Wikinews categories of articles within each clique.

This algorithm satisfies two requirements of a CDEC dataset. First, within each storyline, all articles are related to each other. Second, articles from different storylines aren't adjacent in the Wikinews graph ($W$); thereby, they are very likely unrelated.

For this work, we narrow our focus only to articles in the ``Disaster and Accidents'' category on Wikinews.\footnote{\url{https://en.wikinews.org/wiki/Category:Disasters_and_accidents}} Following the terminology of prior work, our dataset constitutes of a single topic (Disaster and accidents) and 55 subtopics (individual storylines). We restrict CDEC annotations to subtopics that contain 3 or 4 documents. Our algorithm aims for completeness of the CDEC dataset by maximizing for intra-subtopic and minimizing inter-subtopic coreference.
 
\begin{table}[t]
    \centering
    \begin{tabular}{@{}lr@{}}
    \toprule
    \# topics & 1 \\
    \# subtopics & 55 \\
    \# documents & 176 \\
    \# sentences per doc (avg.) & 14.6 \\
    \# tokens per doc (avg.) & 344 \\
    \# event mentions & 7220 \\
    \# mentions per doc (avg.) & 41 \\
    \midrule
    \# document pairs & 198 \\
    \# CDEC links & 4282 \\
    \# CDEC links per document pair & 21.6 \\
    \midrule
    \# full coreference links & 2914 \\
    \# partial coreference links & 1368 \\
    \bottomrule
    \end{tabular}
    \caption{An overview of the compiled CDEC dataset.}
    \label{tab:dataset_statistics}
\end{table}

\paragraph{Event Mention Identification:} To annotate the event mentions in the above-collected documents, we first run a combination of mention detection systems. Specifically, we use the OpenIE system \cite{stanovsky-etal-2018-supervised} from AllenNLP \cite{gardner-etal-2018-allennlp} and an open-domain event extraction system \cite{araki-mitamura-2018-open}. The former is effective at extracting verbal events, whereas the latter is good at nominal events. In contrast to most prior work, we do not restrict the mentions to specific event types or salient events. We believe it is important to study all underlying events to achieve a complete understanding of the corpus. Since the quality of mention identification is critical to our CDEC dataset, we ask an expert to go through the automatically identified mentions and add/edit/delete mentions using the Stave annotation tool \cite{liu-etal-2020-data-centric}.\footnote{the expert annotator is an author of this work.} 

\autoref{tab:dataset_statistics} presents the overall statistics of our document corpus. Our documents are $\sim$14.6 sentences long, comparable to prior work, ECB+ (16.6), GVC (19.2), and FCC (34.4). However, our documents are significantly more dense in terms of event mentions. Our documents contain $\sim$41 mentions (on avg.), much higher compared to prior work, ECB+ (15.3), GVC (14.3), FCC (5.8). Given the dense nature of our documents, we appropriately design our annotation task and interface.

\section{Annotating Coreference via Crowdsourcing}
\label{sec:mturk_setup}

Corefering event mentions share their identity. However, the extent of sharing for them to be considered coreferential is unclear. To empirically study this behavior, we crowdsource annotations on Mechanical Turk. We use the crowd workers' responses to analyze the influence of quasi-identity on coreference decisions.

\subsection{Annotation Task}\label{ssec:annotation_task_definition}

The input to our annotation task constitutes a pair of documents, with all event mentions pre-identified. Annotator iterates through every mention on the left document and select corefering mentions from the right document. We also provide the document titles and publication dates to help set the context for the articles. Note that we focus solely on cross-document coreference in this work and leave the addition of within-document links to future work.

Prior work has highlighted the difficulty in capturing event coreference, specifically in cases where the mentions are only quasi-identical \cite{hovy-etal-2013-events}. Notably, \citet{recasens-etal-2012-annotating} found direct annotation of partial identity to be a difficult task. Therefore, we propose to analyze this behavior by collecting supplementary information from the annotators. For each coreference link created by an annotator, we ask them four \emph{follow-up questions}, 1. overlap in location, 2. overlap in time, 3. overlap in participants, and 4. potential inclusion relationship.\footnote{see \autoref{tab:followup_questions} in Appendix for the exact formulation of these follow-up questions.} Annotators implicitly consider these aspects when making a coreference decision; therefore, responding to these questions won't increase the annotators' cognitive load significantly. As we show in \S\ref{sec:quasi_identity_analysis}, the responses to these questions help us tease apart the cases of partial identity.

Unlike within-document coreference, disjoint narratives between documents often complicate CDEC annotation tasks. \citet{wright-bettner-etal-2019-cross} analyzed this behavior in detail and proposed a new \texttt{contains-subevent} label for within-document links that improved annotator agreement and reduced inconsistencies. However, they rely on experts to create the within-doc \texttt{contains-subevent} label beforehand. Instead, we focus solely on cross-document links and frame the task as a simple pair-wise classification. Our framing allows non-expert annotators to make decisions without concern for complex granularity issues. Our follow-up question regarding inclusion facilitates a post hoc analysis of the event granularities in our dataset.

To ensure completeness of our CDEC dataset, we collect annotations for each pair of documents in a given subtopic (\S\ref{sec:corpus_preparation}). As highlighted earlier, the quasi-identity of events may or may not allow for the application of transitivity property. Therefore, in our dataset, we cannot expand coreference links using transitivity. So collecting annotations between each pair in a given subtopic is necessary.

\paragraph{Annotation Guidelines:} Events are commonplace in the newswire; therefore, it is feasible to explain the concept of events and their coreference via simple example-based guidelines. In our guidelines, we first define \emph{events} and then provide numerous examples of identical and non-identical event mentions, with detailed explanations. Following prior work \cite{song-etal-2018-cross}, we rely on the annotator's intuition to decide coreference.\footnote{see \ref{ssec:annotation_guidelines} in Appendix for complete guidelines.}

\subsection{Annotation Tool}
\label{ssec:annotation_tool}

\begin{figure}[t]
    \centering
    \includegraphics[width=0.48\textwidth]{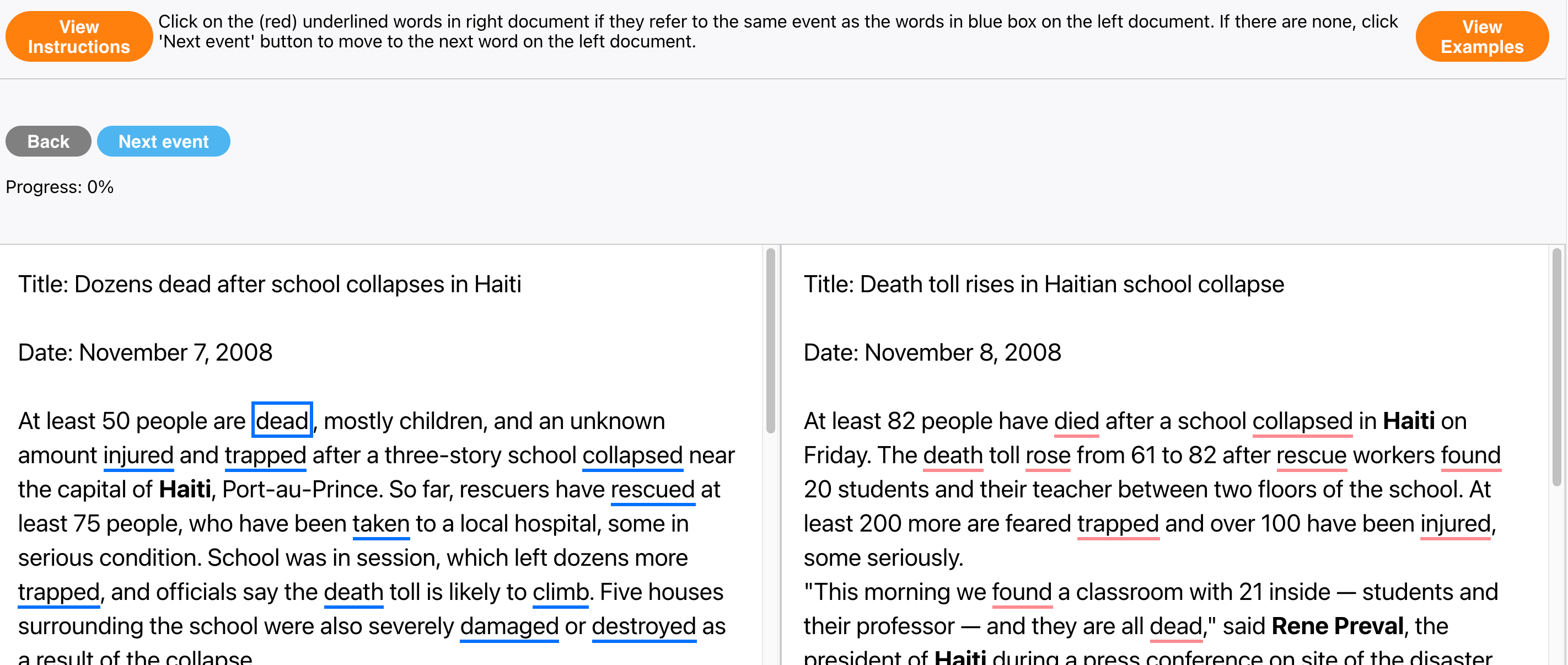}
    \caption{Tool for annotating cross-document event coreference. The two documents are shown side-by-side, with event mentions pre-highlighted. We provide on-screen instructions as well as dedicated pages for viewing detailed instructions and examples. As seen in the example here, we allow annotation of every pair of mentions in the given document pair. In our annotation effort, we present every pair of related documents on this tool, leading to a \emph{densely} annotated dataset.}
    \label{fig:interface_snapshot}
\end{figure}

To efficiently crowdsource annotations, we require a tool that is both easy-to-use and customizable to our workflow. For this purpose, we build upon the Forte\footnote{\url{https://github.com/asyml/forte}} and Stave\footnote{\url{https://github.com/asyml/stave}} toolkits \cite{liu-etal-2020-data-centric}. We extend both the toolkits to support cross-document linking as required by our annotation task. \autoref{fig:interface_snapshot} presents a snapshot of our annotation interface. We highlight event mentions in both the documents and allow the annotator to iterate through each mention on the left document. In addition to dedicated links to instructions and examples, we provide on-screen instructions to assist the annotator in real-time. We also use an English NER tool \cite{ma-hovy-2016-end} to highlight the named entities in the documents. These entities help the annotator keep track of various event participants in the two documents.

We utilize this tool for our entire dataset collection. While we show an application of our annotation tool for CDEC, we believe it's adaptable to other cross-document tasks like entity coreference and event/entity relation labeling tasks. We will release our toolkit to encourage future work on cross-document NLP tasks.

\subsection{Collecting CDEC annotations}
\label{ssec:dataset_collection}

We crowdsource annotations for CDEC using Amazon Mechanical Turk (MTurk). Each Human Intelligence Task (HIT) constitutes annotating cross-document links for one pair of documents. We obtained IRB approval and set our HIT price based on preliminary studies.\footnote{see \ref{ssec:ethical_considerations} in Appendix for more details.} On MTurk, we restricted our HITs to crowd workers from the US and set our qualification thresholds for \% HITs, and total HITs approved as 95\% and 1000 respectively. We paid a fair compensation of \$10.9/hour on average.\footnote{The median pay was slightly higher at \$16.3/hour. Both mean and median pay are above the current minimum wage requirements in the United States.} Our annotation task requires proficiency in English, as well as a good understanding of event coreference. To this end, we attach a qualification test with eight yes/no questions regarding event coreference, with a qualification threshold of 75\%.\footnote{see \ref{ssec:mturk_qualification_test} in Appendix for the test format and the questions.}

For each document pair, we collected annotations from three different crowd workers. In each task, crowd workers go through the two documents and develop a high-level understanding of the news story. They then iterate through the mentions in the left document, in the narrative order, to identify potential cross-document coreference links. From our preliminary studies, we found that annotators spend considerable time reading the two documents. Therefore, to make the best use of the crowd workers' time and effort, we group HITs that constitutes document pairs from the same subtopic. This way, if the crowd worker chooses to, they can annotate the entire subtopic in one sitting, sharing their understanding of a document from one HIT to the next. In total, we collected annotations for 198 document pairs, spanning 176 unique documents and 55 subtopics from 46 crowd workers.

\paragraph{Inter Annotator Agreement (IAA):} For each pair of documents, we collect annotations from three crowd workers. Our setup allows the annotator to decide coreference for every mention pair. To measure IAA, we associate a value to each mention pair (corefering or non-corefering) and compute Krippendorff's $\alpha$. For coreference links, we observed an $\alpha$ of 0.46, indicating moderate agreement \cite{artstein-poesio-2008-survey}.\footnote{It's important to note that we compute IAA on our entire dataset. Our IAA score is comparable to those of quasi-relations from \citet{hovy-etal-2013-events}.} Additionally, we compare the impact of the quasi nature of coreference on the annotator agreement. In our dataset, 31\% of the full-coreference links have a perfect majority (3/3 annotators). However, only 13\% of the partial-coreference links have the same (see \autoref{para:collect_partial_identity} for the methodology used to determine partial coreference). This sharp contrast illustrates the difficulty in capturing partial coreference links.

\paragraph{Selecting CDEC links:} For each pair of mentions, we take a majority vote on the three crowdsourced annotations. In our preliminary analysis, we found many valid coreference links annotated by just one crowd worker. While we encourage the crowd workers to annotate every pair of corefering mentions, they occasionally miss links. Therefore, to ensure completeness of our dataset, we use an adjudicator to go through the single-annotator links to decide if they are in-fact corefering or not.\footnote{the adjudicator is an author of this paper.}

\autoref{tab:dataset_statistics} presents an overview of the compiled CDEC dataset. Unlike prior work, we do not create mention clusters by expanding the links via transitive closure. As we show in \S\ref{sec:quasi_identity_analysis}, quasi-identity of events warrants the need to analyze coreference at the level of mention pairs instead of clusters.
%  (illustrated in \autoref{fig:non_transitivity_example})

\subsection{Dataset Validation}
\label{ssec:dataset_validation}

To facilitate benchmarking future coreference resolution models, we split our dataset into train and test. Of the 55 subtopics, 40 are for model training and development, and 15 are for the unseen test set. Given the importance of the test set quality, we perform expert validation on a randomly selected subset of 18 document pairs from our test set. The expert inspected the annotated coreference links in the subset and found 97.5\% precision (549/563 were corefering). On the other hand, measuring the recall is hard due to a large number of mention pairs. Therefore, we specifically focus on two types of potentially missing coreference links, 1. mention pairs that share the same head lemma (but not annotated as corefering), 2. mention pairs that are part of a non-transitive triplet.\footnote{$(E_A, E_B, E_C)$ is a non-transitive event triplet if $E_A$ corefers with $E_B$, $E_B$ corefers with $E_C$, but $E_A$ and $E_C$ are non-corefering.} Upon inspection by the expert, we find that majority of lemma-match links are non-corefering (50/565 were corefering), while a majority of non-transitive pairs are corefering (149/173 were corefering). This result indicates the scope for improvement in tackling missing coreference links. We leave this extension to future work.

\section{Studying Quasi-Identity of Events}
\label{sec:quasi_identity_analysis}

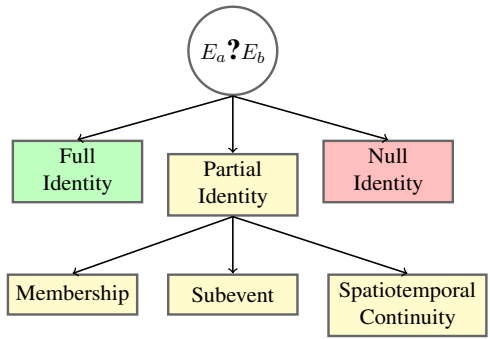
\begin{figure}[t]
\centering
\resizebox{0.4\textwidth}{!}{
\begin{tikzpicture}[
roundnode/.style={circle, draw=black!60, very thick, minimum size=7mm},
rectnode/.style={rectangle, draw=black!60, very thick, minimum size=7mm},
ellipsenode/.style={ellipse, draw=black!60, very thick, minimum size=7mm},
]
\node[roundnode] (mentionpair) {$E_{a}\textbf{\Large{?}}E_{b}$};
\node[rectnode, fill=green!25, text width=2cm, align=center] (full) [below left=1cm and 1cm of mentionpair] {Full\\Identity};
\node[rectnode, fill=yellow!25, text width=2cm, align=center] (partial) [below=1cm of mentionpair] {Partial\\Identity};
\node[rectnode, fill=red!25, text width=2cm, align=center] (null) [below right=1cm and 1cm of mentionpair] {Null\\Identity};
\node[rectnode, fill=yellow!25, text width=2cm, align=center] (membership) [below left=1cm and 0.5cm of partial] {Membership};
\node[rectnode, fill=yellow!25, text width=2cm, align=center] (subevent) [below=1cm of partial] {Subevent};
\node[rectnode, fill=yellow!25, text width=2.5cm, align=center] (spatiotemporal) [below right=1cm and 0.5cm of partial] {Spatiotemporal\\Continuity};

\draw[->,thick] (mentionpair.south) -- (full.north);
\draw[->,thick] (mentionpair.south) -- (partial.north);
\draw[->,thick] (mentionpair.south) -- (null.north);
\draw[->,thick] (partial.south) -- (membership.north);
\draw[->,thick] (partial.south) -- (subevent.north);
\draw[->,thick] (partial.south) -- (spatiotemporal.north);

\end{tikzpicture}}
\caption{A taxonomy of event identity. While full and null identities are well understood, the definition of partial identity is still evolving. We present the three types of partial identity found in our dataset.}
\label{fig:identity_taxonomy}
\end{figure}

% \begin{figure}[t]
% \centering
% \resizebox{0.4\textwidth}{!}{
% \begin{tikzpicture}[
% roundnode/.style={circle, draw=black!60, very thick, minimum size=7mm},
% rectnode/.style={rectangle, draw=black!60, very thick, minimum size=7mm},
% ellipsenode/.style={ellipse, draw=black!60, very thick, minimum size=7mm},
% ]
% \node[roundnode] (mentionpair) {$E_{a}\textbf{\Large{?}}E_{b}$};
% \node[rectnode, fill=green!40, text width=2cm, align=center] (full) [below left=1cm and 1cm of mentionpair] {Full\\Identity};
% \node[rectnode, fill=yellow!40, text width=2cm, align=center] (partial) [below=1cm of mentionpair] {Partial\\Identity};
% \node[rectnode, fill=red!40, text width=2cm, align=center] (null) [below right=1cm and 1cm of mentionpair] {Null\\Identity};
% \node[rectnode, fill=green!10, text width=3cm, align=center] (indistinguishable) [below left=1cm and -0.5cm of partial] {Indistinguishable\\in Context};
% \node[rectnode, fill=red!10, text width=3cm, align=center] (distinguishable) [below right=1cm and -0.5cm of partial] {Distinguishable\\in Context};

% \draw[->,thick] (mentionpair.south) -- (full.north);
% \draw[->,thick] (mentionpair.south) -- (partial.north);
% \draw[->,thick] (mentionpair.south) -- (null.north);
% \draw[->,thick] (partial.south) -- (indistinguishable.north);
% \draw[->,thick] (partial.south) -- (distinguishable.north);

% \end{tikzpicture}}
% \caption{A taxonomy of event identity. While full and null identity are well understood, the definition of partial identity is still evolving.}
% \label{fig:identity_taxonomy}
% \end{figure}
\begin{table*}[ht]
    \centering
    \resizebox{\textwidth}{!}{
    \begin{tabular}{@{}l@{ \ }p{\textwidth}@{}}
    \toprule
    \multicolumn{2}{l}{\emph{Membership}} \\
    \midrule
    1a &  The \textbf{fire} has burned about 4400 acres so far and 15 homes have been lost, however there have been no reported injuries or deaths. \\
    1b & Reports say that the amount of people fleeing from their homes in California located in the United States due to \textbf{wildfires} has reached the 1,000,000 mark as the fires continue to grow. \\
    \midrule
    2a & Several aftershocks have rocked the same area, the latest measuring 7.1, had a depth of 10 km. It was first reported to be a 7.3 \textbf{aftershock}. \\
    2b & Some smaller \textbf{aftershocks} with magnitudes between 5.2 and 5.7 were also reported in the region. \\
    2c & That quake was followed by as many as 60 \textbf{aftershocks} for at least a week, with some ranging as high as magnitude 7.8. \\
    \midrule
    \multicolumn{2}{l}{\emph{Subevent}} \\
    \midrule
    3a & A freight train in Lviv, Ukraine derailed, caught fire, and spilled a toxic chemical, releasing dangerous fumes into the air early Tuesday morning (local time), and people who live near the site of the \textbf{crash} are still becoming sick. \\
    3b & The available information about the phosphorous cloud following the railway \textbf{accident} in the Ukraine last Monday is becoming more and more cryptic. \\
    \midrule
    % \multicolumn{2}{l}{\emph{Subevent}} \\
    % \midrule
    4a & During the fifteen days of the trial, the prosecutors called 92 witnesses to testify as to the chaotic scenes following the \textbf{bombing}. \\
    4b & Two \textbf{explosions} within seconds of each other tore through the finish line at the Boston Marathon, approximately four hours after the start of the men's race. \\
    \midrule
    \multicolumn{2}{l}{\emph{Spatiotemporal Continuity}} \\
    \midrule
    5a & Tropical \textbf{storm} Richard is nearing hurricane strength with winds of 70 mph (115 kph) as it lashes Honduras with heavy rains \\
    5b & \textbf{Hurricane} Richard made landfall in Belize about 20 mi (35 km) south-southeast of Belize City with winds of 90 mph (150 kph) at approximately 6:45 local time (0045 UTC) according to the National Hurricane Center (NHC) \\
    % \midrule
    % 5a & At least 259 people are dead and over 3000 people have been infected in the Haitian cholera \textbf{outbreak}. \\
    % 5b & The Haitian cholera \textbf{outbreak} has killed 292 people and infected over 4000, according to the Haitian government, although there are no new cases in the earthquake ravaged capital, Port-au-Prince. \\
    \bottomrule
    \end{tabular}}
    \caption{An illustration of quasi-identity of event mentions across documents. These examples cover the three identified types of quasi-identity, membership, subevent, and spatiotemporal continuity.}
    \label{tab:quasi_identity_examples}
\end{table*}

Numerous factors determine the identity of an event mention, including the semantics of the mention, arguments (place, time, and participants), and the overall document context. Therefore, overlap in these factors determines the extent of coreference between two given mentions. This overlap leads to cases of partial (quasi-) identity. Our annotation workflow allows for empirical investigation of this phenomenon, and we summarize our observations through a taxonomy of event identity in \autoref{fig:identity_taxonomy}.
Except for \citet{wright-bettner-etal-2019-cross}, prior CDEC datasets do not account for the partial identity during the annotation process. \citet{hovy-etal-2013-events} have previously proposed two types of partial identity, membership, and subevent. In addition to providing evidence for these two types in our dataset, we also identify a novel type of partial identity termed as \emph{spatiotemporal continuity}.

\paragraph{Collecting Partial Identity:}
\label{para:collect_partial_identity}
We use the responses to follow-up questions for qualitatively analyzing cases of partial identity. We consider a link to be a case of partial identity if a strict majority of annotators indicate one of the following. First, there is an inclusion relationship between corefering mentions. Second, the two overlap in place, time, or participants.
With this screening methodology, we found $\sim$32\% of the total CDEC links to be candidates for partial identity (\autoref{tab:dataset_statistics}). We qualitatively analyze the dataset and identify three types of partial identity, 1. Membership, 2. Subevent, and 3. Spatiotemporal continuity. \autoref{tab:quasi_identity_examples} illustrates each type with examples from our compiled dataset.

\paragraph{Membership:} An event mention $E_a$ is a member of event mention $E_b$. Consider the two sentences, 1a, and 1b. The mention `fire' (1a) denotes a specific wildfire, whereas `wildfires' (1b) denotes a group of wildfires, including the one in 1a. The concept of partial identity often challenges the transitivity assumption of coreference. For instance, the mentions [smaller] `aftershocks' (2b) and [7.1] `aftershock' (2a) share no identity, thereby, non-coreferential. However, both the mentions partially corefer with [60] `aftershocks' from 2c.

\paragraph{Subevent:} An event mention $E_a$ is a subevent of event mention $E_b$. This behavior can be seen in the coreference between the `crash' event from 3a, and the `accident' event from 3b. While the `accident' event involves many individual events, derailed, caught fire, spill chemical, and release fumes, it partially corefers with the event `crash' from 3a that likely refers only to the derailment. Similarly, consider the case of the Boston Marathon Bombing in examples 4a and 4b. The `bombing' event from 4a refers to the whole incident, whereas the `explosions' in 4b refers to specific subevents of the `bombing'.

\paragraph{Spatiotemporal Continuity:} The identity of an event can \textit{continuously} evolve over space and time. Consider the two mentions, `storm' and `Hurricane' from \autoref{tab:quasi_identity_examples} (5a, 5b). At a high level, these mentions are corefering because they denote the same event (storm Richard). However, the expressions of this event differ slightly across the two documents. In the former, it's a storm (with 70mph winds) having an impact in Honduras, whereas, in the latter, it's a hurricane (with 90mph winds) impacting Belize. Similar behavior is visible with the [Haitian cholera] `outbreak' event from \autoref{fig:spatiotemporal_continuity_example}. The outbreak gradually evolves, with growing infection (2600 $\rightarrow$ 3000 $\rightarrow$ 4000) and deaths (200 $\rightarrow$ 259 $\rightarrow$ 292). In both of these examples, we observe the event changes gradually and is always continuous in both space and time dimensions.\footnote{We borrow the term spatiotemporal continuity from the Philosophy literature. It describes the properties of well-behaved objects \cite{wiggins1967identity}. A similar treatment for entities is presented in \citet{Recasens2011IdentityNA}.}

In line with prior work on entities \cite{Recasens2011IdentityNA}, we believe identity and coreference of events to be a continuum. Our dataset already includes many instances of partial identity to support this hypothesis. The above-described cases of partial identity (membership, subevent, and spatiotemporal continuity) will pose new challenges to future dataset collection efforts. We believe our annotation workflow and guidelines will be of use to future work.

In this section, we establish a clear case for tackling partial identity within the coreference resolution task. However, in practical settings, the boundaries between full, partial, and null identities remain fuzzy. As seen in our analysis on the inter-annotator agreement, humans find it hard to identify cases of partial coreference. In the downstream coreference resolution task, users are primarily interested in knowing if two given mentions share an identity or not. Therefore, we propose to view both full and partial identity under a single coreference label (`coreference') and contrast them against cases with no shared identity (`non-coreference'). Compared to prior datasets, this presents new challenges in tackling partial identity within the `coreference' label. 

\section{Baselines}
\label{sec:baselines}

We define the task as a mention pair classification problem. Due to the quasi-identity nature of event mentions (\S\ref{sec:quasi_identity_analysis}), we do not cluster mentions in coreference groups. Additionally, we consider both full and partial identity under the coreference label. We present two baseline models, lemma-match, and a cross-encoder model. We split the dataset of 55 subtopics into train and test, with 40 subtopics for training and development, and 15 subtopics for the held-out test set. For our experiments, we assume gold mentions and subtopic information.\footnote{topic-level performance \cite{cattan-etal-2021-realistic}}

\paragraph{Lemma-match:} For our first baseline, we implement the traditional lemma-match baseline. We use spacy's large model\footnote{\texttt{en\_core\_web\_lg} from \url{https://spacy.io}} to extract the head lemma of the event mentions, and consider two mentions corefering if the lemma's match. Following \citet{upadhyay-etal-2016-revisiting}, we also experiment with a Lemma-$\delta$ baseline. In our experiments, we found the best dev performance with $\delta$=0, resolving to a simple lemma baseline. This could be due to our assumption of access to gold subtopic information.

\paragraph{Cross-Encoder:} As a second baseline, we implement BERT-based cross-encoder model. The input consists of a pair of sentences with both mentions highlighted using special tokens to indicate the start and end of mention spans (<E>, </E>). We first concatenate the two event-tagged sentences (with [SEP] token) and pass it through a bert-base-uncased encoder. We then perform mean pooling on the event start tags (<E>), and pass the pooled embedding through a linear classification layer to predict coreference vs. non-coreference. For training the cross-encoder, in addition to the positive coreference pairs, we generate two types of negative mention pairs. For the first type, we collect non-coreference mention pairs from sentences that have a coreference link between a different mention pair. For the second type, we extract non-coreference mention pairs from random sentence pairs between the documents. During training, we use a dataset ratio of 1:5:5 (positive:negative-I:negative-II). We use huggingface transformers \cite{wolf-etal-2020-transformers}, and train the model using AdamW \cite{loshchilov2018decoupled} with an initial learning rate of 2e-5. We also use a linear warmup scheduler, with 10\% of training steps for warmup. We finetune the \# epochs and positive:negative dataset ratio during the development stage (5-fold cross-validation) and use the best configuration when training on the entire train set.

\paragraph{Results:} \autoref{tab:baseline_results} presents the results of our baselines. For model development, we perform 5-fold cross-validation on the training set (40 subtopics). To report the results on the held-out test set (15 subtopics), we train the model's best configuration on the entire training set. We report precision, recall, and F1 scores of the coreference label averaged on five different runs. The lemma baseline only achieves an F1 score of 48.2, indicating that the proposed dataset is lexically diverse. The cross-encoder improves upon the lemma baseline, especially on the recall. Upon inspection of development set predictions, we observe two possible error cases for the cross-encoder model. First, the model struggles at the cases of partial identity (`explosion' vs. `incident' and `evacuate' vs. `evacuations'). This drawback of cross-encoder indicates that the model requires a deeper understanding of event identity. Second, the cross-encoder model is often limited by the information available in a single sentence. It is known the event arguments are often underspecified in the local context \cite{ebner-etal-2020-multi}; therefore, increasing the context to a paragraph or the entire document might help improve the performance.

\begin{table}[t]
    \centering
    \resizebox{0.48\textwidth}{!}{
    \begin{tabular}{@{}l@{ \ }c@{ \ }c@{ \ }c@{ \ }c@{ \ }c@{ \ }c@{}}
    \toprule
    \textbf{Model} & \multicolumn{3}{c}{\textbf{Dev}} & \multicolumn{3}{c}{\textbf{Test}} \\
    & P & R & F1 & P & R & F1 \\
    \midrule
    Lemma-match & 46.6 & 54.9 & 49.9 & 42.3 & 56.0 & 48.2 \\
    \midrule
    Cross-Encoder & 43.1 & 75.4 & 54.3 & 45.9 & 77.3 & 57.6 \\
    & \small{$\pm$ 0.6} & \small{$\pm$ 0.5} & \small{$\pm$ 0.5} & \small{$\pm$ 0.8} & \small{$\pm$ 1.1} & \small{$\pm$ 0.6} \\
    \bottomrule
    \end{tabular}}
    \caption{Baseline results on development and test sets. For cross-encoder, we report the average scores and their standard deviation across five runs.}
    \label{tab:baseline_results}
\end{table}

\section{Conclusion \& Future Work}
\label{sec:conclusion}

In this work, we present a study of the identity of events through annotation of cross-document event coreference. We use a custom-designed annotation tool to collect coreference annotations on a subset of English Wikinews articles. We release our dataset, CDEC-WN, under an open license to encourage further research on event coreference. By collecting evidence for the extent of shared identity between events, we identify three types of partial-identity, membership, subevent, and spatiotemporal continuity. To serve as a benchmark for future coreference resolution systems, we provide results on two baseline models, lemma-match and BERT-based cross-encoder. We believe that our work will encourage further research on the identity of events in the context of CDEC. Potential future directions include expanding CDEC-WN to include within-document coreference links, designing coreference resolution systems that account for cases of partial identity between mentions, and expanding the study of the partial identity of event coreference to new domains.

\section*{Acknowledgements}

We thank the anonymous reviwers for their valuable feedback. We also thank the Mechanical Turk workers for their help in our annotation process. This material is based on research sponsored by the Air Force Research Laboratory under agreement number FA8750-19-2-0200. The U.S. Government is authorized to reproduce and distribute reprints for Governmental purposes notwithstanding any copyright notation thereon. The views and conclusions contained herein are those of the authors and should not be interpreted as necessarily representing the official policies or endorsements, either expressed or implied, of the Air Force Research Laboratory or the U.S. Government. 
% \hector{remember to get the funding project number from teruko}

\bibliography{anthology,custom}
\bibliographystyle{acl_natbib}

\clearpage
\newpage
\appendix

\section{Appendix}

\subsection{Ethical Considerations}
\label{ssec:ethical_considerations}
% this is required according to EMNLP guidelines, refer to the CFP for more information, https://2021.emnlp.org/call-for-papers/ethics-faq#what-should-i-be-aware-of-if-my-paper-presents-a-new-dataset

In our dataset construction, we follow the standard norms for ethical research involving human participants. We obtained IRB approval before starting our study. Our pilot study indicated that each HIT takes $\sim$10-15 minutes; therefore, we set the price of individual HIT to be \$2.3. Overall, we paid a fair compensation of \$10.9/hour (with median pay of \$16.3/hour). For each HIT, the crowd workers on Mechanical Turk have signed the informed consent form before starting the task (see \ref{ssec:mturk_consent_form} in Appendix). We provided clear instructions for using our annotation tool, both within and through an instructional video. We provide positive and negative examples to illustrate event coreference to the crowd workers (see \ref{ssec:annotation_guidelines} in Appendix). Our dataset is limited to the English language, specifically for text documents relating to Disasters and accidents. While we have taken specific steps to improve the quality of our dataset, there might be incorrect or missing coreference links. However, we believe that such incorrect/missing links will not create additional risks to the models trained on our dataset.

\subsection{Annotation Guidelines}
\label{ssec:annotation_guidelines}

To explain the task of cross-document event coreference to crowd workers on Mechanical Turk, we present detailed example-based guidelines (\autoref{tab:interface_examples_1}, \autoref{tab:interface_examples_2}). Additionally, we provide crowd workers with detailed instructions to our annotation interface (\autoref{tab:interface_instructions_1}, \autoref{tab:interface_instructions_2}). Workers view these instructions before the start of each task and optionally during the task. In our HIT, we also link to a 1-minute video tour of our annotation interface.

\begin{table*}[t]
\centering
\resizebox{\textwidth}{!}{
\begin{tabular}{p{\textwidth}}
\toprule
\large{\textbf{Instructions for using the tool}} \\
\\
This tool can be used to select events that are the same across the two given documents. \\
\\
\textbf{How to open instructions} \\
<embedded GIF> \\
\textbf{How to annotate one pair of events} \\
<embedded GIF> \\
\textbf{How to delete previous annotations} \\
<embedded GIF> \\
\textbf{How to proceed to the next event} \\
<embedded GIF> \\
\\
At any point during the task, you can click on the ``View Instructions'' button to read these instructions. \\
\\
\textbf{What is this task about?}
\begin{itemize}
    \item Two related documents are presented side-by-side on the tool.
    \item A few words in both the documents are underlined and these are referred to as events.
    \item The task is to select events from the right document that are the same as the currently highlighted event in the left document.
\end{itemize}
\textbf{How should I solve this task?}
\begin{itemize}
    \item When you first start the task, make sure you read through both the left and right documents to get an overall understanding of the two documents.
    \item At each step, an event is highlighted in a blue box on the left document (aka. target event). Now, your goal is to identify underlined events from the right document that are the same as the target event from the left document.
    \item Once you select an event from the right document (an annotation), you are presented a few follow-up questions. Make sure you answer these questions to the best of your knowledge.
    \item If you change your mind while answering the questions, you can click the ``Cancel'' button to remove your annotation.
    \item After you have identified all possible same events from the right document (if any), please use the ``Next event'' button to move to the next target event on the left document.
\end{itemize}
\\
\bottomrule
\end{tabular}}
\caption{Instructions as shown to the annotators on the interface.}
\label{tab:interface_instructions_1}
\end{table*}

\begin{table*}[t]
\centering
\resizebox{\textwidth}{!}{
\begin{tabular}{p{\textwidth}}
\toprule
\large{\textbf{Instructions for using the tool} (contd.)} \\
\\
\textbf{FAQs} \\
\\
\textbf{Q:} I made a mistake and incorrectly marked two events as the same. How do I correct this? \\
If you are still answering the follow-up questions, you can just click on the ``Cancel'' button. If you have already moved to the next target event, you can use the ``Back'' button to move back the previously finished target events. \\
\\
\textbf{Q:} I am not sure how to respond to the follow-up questions. How should I proceed? \\
The follow-up questions help us understand more about your decision that two events are the same. It is important to note that the response to these questions need not always be ``Yes''. In fact, in many cases, you may not have enough information to respond with a definite ``Yes'' or ``No'', then please feel free to select ``Not enough information''. \\
\\
\textbf{Q:} How do I decide if two events are the same or different? \\
We understand that this decision is not always easy. To help you with this, we compiled a bunch of examples. You can quickly glance through them using the ``View Examples'' button on the tool. \\
\\
\textbf{Q:} How do I contact the authors of the task? \\
\indent For any comments, feedback and/or suggestions, please use this form (XXXX). We strive to make this a great experience for you.
\\
\bottomrule
\end{tabular}}
\caption{Instructions as shown to the annotators on the interface. (contd)}
\label{tab:interface_instructions_2}
\end{table*}
\begin{table*}[t]
\centering
\resizebox{\textwidth}{!}{
\begin{tabular}{p{\textwidth}}
\toprule
\large{\textbf{Examples}} \\
\\
\textbf{Goal of the Task} \\
You will help us identify the same events from different documents. \\
\\
\textbf{What is an event?} \\
People use text to describe what happen(ed) in the world. These are called events in text. We often use verbs, sometimes even (pro)nouns, and adjectives as \underline{events}. For example: \\
\\
It \underline{rained} a lot yesterday. \\
There was a \underline{fire} last night. \\
He \underline{got sick}. \\
\\
\textbf{How do we know that the two events are the same?} \\
In the following examples (1 to 5), two events are the same. \\
\begin{description}
\item[1.] When two events refer to the same thing, they should be the same in terms of meaning, or semantically identical.
\begin{itemize}
    \item Taken as a whole, the evidence suggests that the plan to \underline{bomb} the Boston Marathon took shape over three months.
    \item Dzhokhar Tsarnaev apologized for suffering caused by the Boston Marathon \underline{bombing}.
\end{itemize}
\item[2.] When two events are the same, one event may be the synonym for the other.
\begin{itemize}
    \item A 16-year-old southern Utah boy was \underline{accused} of bringing a homemade bomb to his high school.
    \item The teen was \underline{charged} Monday with attempted murder and use of a weapon of mass desctuction, both first-degreen felonies.
\end{itemize}
\item[3.] Sometimes one event may be the pronoun (e.g.,it) or the anaphora (e.g., this, that) of the other, when they are the same.
\begin{itemize}
    \item Both drones carried explosives, and no YPF (``People's Defence Units'') fighters were injured in the \underline{incident}.
    \item \underline{This} would not be the first terrorist drone strike.
\end{itemize}
\item[4.] The same events do not have to take place at the same time. In the following example, one event (``go'') would happen in the future, while the other (``went'') did occur.
\begin{itemize}
    \item The couple had been planning to \underline{go} to Paris for a long time.
    \item They finally \underline{went} there last month.
\end{itemize}
\item[5.] Sometimes the same events are described from different perspectives. The following example refers to the exchange of the gift from two perspectives.
\begin{itemize}
    \item John \underline{gave} a gift to Mary.
    \item Mary \underline{received} a gift from John.
\end{itemize}
\end{description}
\\
\bottomrule
\end{tabular}}
\caption{Examples for coreference and non-coreference, as shown to the annotators on the interface.}
\label{tab:interface_examples_1}
\end{table*}

\begin{table*}[t]
\centering
\resizebox{\textwidth}{!}{
\begin{tabular}{p{\textwidth}}
\toprule
\large{\textbf{Examples} (contd.)} \\
\\
In the following examples (6 to 8), two events are not the same.

\begin{description}
\item[6.] When one event is a part of the other larger event, they are not the same.
\begin{itemize}
    \item Following the trial of Mahammed Alameh, the first suspect in the \underline{bombing}, investigators discovered a jumble of chemicals, chemistry implements and detonating materials.
    \item The \underline{explosion} killed at least five people. (``bombing'' refers to the entire process which starts with making a bomb and ends with destructions, damages and injuries, while ``explosion'' is a smaller event that occurs in that processes)
\end{itemize}
\item[7.] Two events are not the same even if they are the same semantically. The first example refers to the general bomb-making process, while the second one indicates a particular bomb-making event that took place in the garage.
\begin{itemize}
    \item They obtained the online manual of bomb-making. (general bomb-\underline{making} process)
    \item They \underline{made} a bomb in the garage. (specific bomb-making event that happened in the specific place)
\end{itemize}
\item[8.] When one event consists of, or is a member of the other event, they are not the same. The first example refers to the specific death of a 44-year-old man, while the second one refers to the deaths of 305 people.
\begin{itemize}
    \item The government announced that a 44-year-old man \underline{died} from the COVID. (death of a 44-year-old man)
    \item There are more than 14,300 confirmed COVID cases, and 305 people have \underline{died}. (deaths of 305 people)
\end{itemize}
\end{description}
\\
\bottomrule
\end{tabular}}
\caption{Examples for coreference and non-coreference, as shown to the annotators on the interface. (contd)}
\label{tab:interface_examples_2}
\end{table*}

In our guidelines, we only present examples of full and null coreference. While we consider membership a form of coreference (partial), we don't train the crowd workers on full and partial identity.

\subsection{MTurk Consent Form}
\label{ssec:mturk_consent_form}

A consent form is attached to the start of each HIT. Crowd workers are required to go through the form and provide their consent before starting the task. Anonymized version of the consent form is presented in \autoref{tab:consent_form_1} and \autoref{tab:consent_form_2}. We anonymize the document for the conference review process.

\begin{table*}[t]
\centering
\resizebox{\textwidth}{!}{
\begin{tabular}{p{\textwidth}}
\toprule
\large{\textbf{Consent From}} \\
\\
This task is part of a research study conducted by XXX at XXX and is funded by XXX. \\
\\
\textbf{Purpose} \\
The goal of this study is to collect datasets of coreference-labeled pairs sampled from public online news articles through the help of crowd workers. \\
\\
\textbf{Procedures} \\
You will be directed to a website implemented by the research team to complete the task. You will be asked to read upto 3 pairs of articles. For each pair of articles, you will need to label pieces of text that refer to the same event, and answer additional questions about your labeling. Labeling one pair of articles whose length sums up to 40 sentences is expected to take around 15 minutes. \\
\\
\textbf{Participant Requirements} \\
Participation in this study is limited to individuals age 18 and older, and native English speakers. \\
\\
\textbf{Risks} \\
The risks and discomfort associated with participation in this study are no greater than those ordinarily encountered in daily life or during other online activities. \\
\\
\textbf{Benefits} \\
There may be no personal benefit from your participation in the study but the knowledge received may be of value to humanity. \\
\\
\textbf{Compensation \& Costs} \\
For this task, you will receive between \$2 to \$3 for annotating each pair of articles. The exact reward for each pair depends on the length of corresponding articles. You will not be compensated if you provide annotations of poor quality. \\
There will be no cost to you if you participate in this study. \\
\\
\textbf{Future Use of Information and/or Bio-Specimens} \\
In the future, once we have removed all identifiable information from your data (information or bio-specimens), we may use the data for our future research studies, or we may distribute the data to other researchers for their research studies. We would do this without getting additional informed consent from you (or your legally authorized representative). Sharing of data with other researchers will only be done in such a manner that you will not be identified. \\
\\
\textbf{Confidentiality} \\
The data captured for the research does not include any personally identifiable information about you except your IP address and Mechanical Turk worker ID. \\
By participating in this research, you understand and agree that XXX may be required to disclose your consent form, data and other personally identifiable information as required by law, regulation, subpoena or court order. Otherwise, your confidentiality will be maintained in the following manner: \\
\bottomrule
\end{tabular}}
\caption{Consent Form attached to each of our HITs. We anonymize the document for the conference review process.}
\label{tab:consent_form_1}

\end{table*}

\begin{table*}[t]
\centering
\resizebox{\textwidth}{!}{
\begin{tabular}{p{\textwidth}}
\toprule
\large{\textbf{Consent From (contd.)}} \\
\\
\textbf{Confidentiality (contd.)} \\
Your data and consent form will be kept separate. Your consent form will be stored in a secure location on XXX property and will not be disclosed to third parties. By participating, you understand and agree that the data and information gathered during this study may be used by XXX and published and/or disclosed by XXX to others outside of XXX. However, your name, address, contact information and other direct personal identifiers will not be mentioned in any such publication or dissemination of the research data and/or results by XXX. Note that per regulation all research data must be kept for a minimum of 3 years. \\
The Federal government offices that oversee the protection of human subjects in research will have access to research records to ensure protection of research subjects. \\
\\
\textbf{Right to Ask Questions \& Contact Information} \\
If you have any questions about this study, you should feel free to ask them by contacting the Principal Investigator now at XXX, XXX, or by phone at XXX, or via email at XXX. If you have questions later, desire additional information, or wish to withdraw your participation please contact the Principal Investigator by mail, phone or e-mail in accordance with the contact information listed above. \\
If you have questions pertaining to your rights as a research participant; or to report concerns to this study, you should contact the XXX at XXX. Email: XXX. Phone: XXX or XXX. \\
\\
\textbf{Voluntary Participation} \\
Your participation in this research is voluntary. You may discontinue participation at any time during the research activity. You may print a copy of this consent form for your records. \\
\\
I am age 18 or older. \framebox(7,7){} Yes \framebox(7,7){} No \\
\\
I have read and understand the information above. \framebox(7,7){} Yes  \framebox(7,7){} No \\
\\
I want to participate in this research and continue with the task. \framebox(7,7){} Yes \framebox(7,7){} No \\
\bottomrule
\end{tabular}}
\caption{Consent Form attached to each of our HITs. We anonymize the document for the conference review process. (contd)}
\label{tab:consent_form_2}
\end{table*}

\subsection{MTurk Qualification Test}
\label{ssec:mturk_qualification_test}

To identify high-quality crowd workers, we design a qualification test and add it as an additional requirement to solving our HITs.

\subsubsection{Test Questions}
\label{sssec:test_questions}

In the qualification test on MTurk, we randomly select eight questions from a pool of 20 questions. \autoref{tab:qualification_test_1} and \autoref{tab:qualification_test_2} list all the questions.

\begin{table*}[t]
    \centering
    \resizebox{\textwidth}{!}{
    \begin{tabular}{@{}lp{0.75\textwidth}cc@{}}
    \toprule
    \# & Text & Answer & Type \\
    \midrule
    1 & A 500lb bomb packed in the Cavalier is \underline{detonated} with a remote trigger. The \underline{explosion} tears through Market Street. & yes & Synonym \\
    2 & The \underline{death} toll of the Omagh bomb blast in Northern Ireland has risen to 29 following the \underline{death} of a man in hospital. & no & Member \\
    3 & Ahmed al-Mughassil was \underline{arrested} in Beirut and transferred to Riyadh, the Saudi capital, according to the Saudi newspaper Asharq Alawsat. The Saudi Interior Ministry and Lebanese authorities had no immediate comment on the \underline{capture}. & yes & Synonym \\
    4 & The blast didn’t cause the \underline{destruction} its planners intended. But it \underline{opened up} a multi-story crater in the building, injured more than 1,000 people and ultimately killed six. & no & Member \\
    5 & March 4, 1998 - Four defendants, Salameh, Ayyad, Abouhalima, and Ajaj, are convicted. They are \underline{sentenced} to prison terms of 240 years each. In 1998, the sentences were vacated. In 1999, the men were \underline{re-sentenced} to terms of more than 100 years. & no & Unrelated \\
    6 & Perhaps the only early clues to emerge on an early quiet second day of the Boston Marathon bombing \underline{investigation} - from the ATF and the FBI and the Boston police, from anonymous law enforcement officials and doctors pulling ball bearings out of victims limbs - concern the Boston bombs themselves. A similar scene played out in the Boston suburb of Newton, where a bomb used a robot to \underline{investigate} a suspicious object that turned out to be a circuit board. & no & Member \\
    7 & As of Tuesday morning, jurors began reviewing evidence and witness testimony, which will play a role in helping them divide Dzhokhar Tsarnaev’s \underline{guilt} on each of the 30 charges he faces. A key issue for jurors - both in the guilt phase and later the penalty phase if Tsarnaev is \underline{convicted} - will be whether the jurors see Tsarnaev as an equal partner with his old brother, Tamerian Tsarnaev, in the Boston Marathon bombing and the violent events that followed. & yes & Synonym \\
    8 & Though \underline{building} the bomb was relatively easy, the experts say, it was not by any means free of danger. The bulkiest part of the bomb, they say, was extremely stable and could only have been touched off with a tremendous kick, like that provided by nitroglycerine. Making the nitroglycerin, blending some of the chemicals, was the trickiest part of the \underline{process}. & yes & Synonym \\
    9 & An ongoing Somali \underline{offensive}, backed by the U.S. and an African Union peacekeeping force has recaptured territory from al Shabaab in south-central Somalia, but has not eliminated al Shabaab’s ability to conduct VBIED attacks. U.S.-backed Somali ground \underline{operations} along with improved counter-VBIED capabilities among Somali forces may have slightly decreased VBIED attacks between November 2017 and January 2018. & yes & Synonym \\
    10 & According to the United Nations, more than 2.3 million Venezuelans have left their country in recent years. Increasingly they are leaving with no money and are \underline{traveling} on foot across South American countries like Colombia, Ecuador and Peru, in dangerous \underline{journeys} that can take several weeks. & no & Member \\
    \bottomrule
    \end{tabular}}
    \caption{Examples used with the qualification test on Mechanical Turk. For each paragraph with two highlighted events, we ask the question, ``In the above paragraph, are the highlighted events the same?". The crowd worker has to select one of the ``Yes" or ``No" options.}
    \label{tab:qualification_test_1}
\end{table*}

\begin{table*}[t]
    \centering
    \resizebox{\textwidth}{!}{
    \begin{tabular}{@{}lp{0.75\textwidth}cc@{}}
    \toprule
    \# & Text & Answer & Type \\
    \midrule
    11 & Spain’s King Juan Coarlos and Queen Sofia traveled to their summer residence in Majorca Saturday just two days after a \underline{bombing} blamed on Basuqe separatists \underline{killed} two policemen on the resort island. & no & Member \\
    12 & Yahoo Inc. is preparing to \underline{lay off} between 600 and 700 workers in the latest shakeup triggered by the Internet company lackluster growth. Employees could be notified of the \underline{job cuts} as early as Tuesday, according to a person familiar with  Yahoo’s plans. & yes & Synonym \\
    13 & A man shot and killed by police officers during a burglary here early Monday was identified by law enforcement authorities as the suspect in a string of five shooting \underline{deaths} in South Carolina over the last 10 days. Sheriff Bill Blanton of Cherokee Country, S.C., where the \underline{killings} took place, confirmed Monday evening that the authorities had been seeking the man killed in the burglary, Patrick T. Burris, a felon with a long record who had served seven years in prison and was paroled in April. & yes & Synonym \\
    14 & Staff Sgt. Robert Bales offered a tearful \underline{apology} Thursday for gunning down 16 unarmed Afghan civilians inside their homes but said he still could not explain why he had carried out one of the worst U.S. war crimes in years. The unsworn \underline{statement} from Bales, 40, came on the third day of hearing to determine whether he should ever be eligible for parole in the March 2012 Massacre.  & yes & Synonym \\
    15 & In January two men were \underline{hanged} after being convicted of involvement in protests, and in May, four Iranian Kurds and another man accused of terrorism were \underline{executed}.  & no & Unrelated \\
    16 & The Dow Corning Corporation filed for \underline{bankruptcy} protection in a Federal court in Bay City, Michigan. Dow Corning said that seeking the protection of the \underline{bankruptcy} court was the only way it could devise an enforceable plan to deal with the claims against it. & no & Realis \\
    17 & The UN report accused both Israel and Palestinian armed groups of commiting \underline{war} crimes during the three-week \underline{war} in Gaza that erupted on December 27, killing some 1,400 Palestinians and 13 Israelis. & no & Realis \\
    18 & A judge has ordered the surviving children of the Rev. Martin Luther King Jr. and Coretta Scott King to hold a shareholder’s \underline{meeting} to discuss their father’s estate. The three siblings are the sole shareholders, directors and officers of a company that manages their father’s intellectual property, but they have not \underline{met} for an annual shareholder’s meeting since 2004. & no & Realis \\
    19 & The first \underline{attack} was a failure, but if the report is accurate, then it signals a dangerous new terror threat. The report showed pictures of the remains of a homemade \underline{attack} drone. & no & Realis \\
    20 & A key issue for jurors - both in the guilt phase and later in the penalty phase if Tsarnaev is convicted - will be whether the jurors see Tsarnaev as an equal partner with his older brother, Tamerlan Tsarnaev, in the Boston Marathon \underline{bombing} and the violent events that followed. Taken as a whole, the evidence suggests that the plan to \underline{bomb} the Boston Marathon took shape over three months. & yes & Realis \\
    \bottomrule
    \end{tabular}}
    \caption{Examples used with the qualification test on Mechanical Turk. For each paragraph with two highlighted events, we ask the question, ``In the above paragraph, are the highlighted events the same?". The crowd worker has to select one of the ``Yes" or ``No" options. (contd)}
    \label{tab:qualification_test_2}
\end{table*}

\subsubsection{Test Format}
\label{sssec:test_format}

\autoref{tab:example_qualification_test} presents the format of the qualification test used for screening crowd workers.

\begin{table*}[t]
\centering
\resizebox{\textwidth}{!}{
\begin{tabular}{p{\textwidth}}
\toprule
\large{\textbf{Screening Test}} \\
\\
In this test, we ask you to identify whether two events (\textbf{highlighted} in each paragraph) indicate the same thing or not. Read each paragraph carefully and answer the question by selecting the appropriate option, \textit{Yes} or \textit{No}. \\
\\
In total, you are presented with 8 questions and the time limit for this test is 20 minutes. \\
\\
\textbf{Note}: It is important you do this test on your own because our HITs are similar to the questions presented in this test. For your reference, we provide five examples below, \\
\\
He \textbf{died} of injuries from the accident. His friends were all saddened to hear his \textbf{death}. \\
\textit{Question}: In the above paragraph, are the highlighted events the same? \\
\textit{Answer}: Yes (both words, \textbf{died} and \textbf{death} indicate the person's death) \\
\\
The suspect was \textbf{shot} and killed in the \textbf{raid} by the armed officers. \\
\textit{Question}: In the above paragraph, are the highlighted events the same? \\
\textit{Answer}: No (\textbf{shot} happened during the \textbf{raid}) \\
\\
The couple had been planning to \textbf{go} to Paris for a long time. They finally \textbf{went} there last month. \\
\textit{Question}: In the above paragraph, are the highlighted events the same? \\
\textit{Answer}: Yes (The two events do not have to take place at the same time. Here, \textbf{go} would happen in the future, and \textbf{went} did occur.) \\
\\
John \textbf{gave} a gift to Mary. Mary \textbf{received} a gift from John. \\
\textit{Question}: In the above paragraph, are the highlighted events the same? \\
\textit{Answer}: Yes (Same events described from different perspectives.) \\
\\
Following  the trial of Mahammed Alameh, the first suspect in the \textbf{bombing}, investigators discovered a jumble of chemicals, chemistry implements and detonating materials. The \textbf{explosion} killed at least five people. \\
\textit{Question}: In the above paragraph, are the highlighted events the same? \\
\textit{Answer}: No (One event is part of the other larger event. \textbf{bombing} refers to the entire process which starts with making a bomb and ends with destructions, damages and injuries, while \textbf{explosion} is a smaller event that occurs in that processes.) \\
\\
Q1. .... \\
\framebox(7,7){} Yes \ \ \framebox(7,7){} No \\
Q2. .... \\
\framebox(7,7){} Yes \ \ \framebox(7,7){} No \\
... \\
\bottomrule
\end{tabular}}
\caption{The template used in the qualification test to screen annotators. In addition to instructions and examples, we present eight yes/no questions.}
\label{tab:example_qualification_test}
\end{table*}

\subsection{HIT Template}
\label{ssec:hit_template}

\autoref{tab:hit_template} presents our HIT layout. Our layout is simple, and all of our annotations are collected using our custom-designed annotation tool.

\begin{table*}[t]
\centering
\resizebox{\textwidth}{!}{
\begin{tabular}{p{\textwidth}}
\toprule
\large{\textbf{Annotating Event Coreference in News Articles}} \\
\\
In this HIT, you will be using our tool to perform the task. For a short tutorial on using our interface, see this 1 minute video: XXX. This HIT contains the following two steps,
\begin{itemize}
    \item Visit the URL provided below to perform the task.
    \item At the end of the task, you will be provided a secret code. To submit this HIT, copy the secret code and paste it into the box provided below. Note that the secret code is unique for each task.
\end{itemize}
Link to the task: XXX \\
\\
\textbf{Fill in the secret code} \\
\\
Paste the secret code provided at the end of the task into the text box (*required) \\
\\
\framebox(300, 50){} \\
\bottomrule
\end{tabular}}
\caption{The template used for each Human Intelligence Task (HIT) on Mechanical Turk.}
\label{tab:hit_template}
\end{table*}

\subsection{Follow-up Questions}
\label{ssec:followup_questions}

\autoref{tab:followup_questions} lists the four follow-up questions. We present these questions for each coreference link annotated by the crowd worker.

\begin{table*}[t]
\centering
\resizebox{\textwidth}{!}{
\begin{tabular}{p{\textwidth}}
\toprule
\textbf{Place}: Do you think the two events happen at the same place? \\ \\
\framebox(7,7){} \ Exactly the same \hfill \framebox(7,7){} \ The places overlap \hfill \framebox(7,7){} \ Not at all \hfill \framebox(7,7){} \ Cannot determine \\ \\
\textbf{Time}: Do you think the two events happen at the same time? \\ \\
\framebox(7,7){} \ Exactly the same \hfill \framebox(7,7){} \ They overlap in time \hfill \framebox(7,7){} \ Not at all \hfill \framebox(7,7){} \ Cannot determine \\ \\
\textbf{Participants}: Do you think the two events have the same participants? \\ \\
\framebox(7,7){} \ Exactly the same \hfill \framebox(7,7){} \ They share some participants \hfill \framebox(7,7){} \ Not at all \hfill \framebox(7,7){} \ Cannot determine \\ \\
\textbf{Inclusion}: Do you think one of the events is part of the other? \\ \\
\framebox(7,7){} \ Yes, the left event is part of right one \hfill \framebox(7,7){} \ Yes, the right event is part of left one \\ \framebox(7,7){} \ No, they are exactly the same \hfill \framebox(7,7){} \ Cannot determine \\ \\
\bottomrule
\end{tabular}}
\caption{Follow-up questions used for each annotated coreference link.}
\label{tab:followup_questions}
\end{table*}

\end{document}